\documentclass{article}
\usepackage[utf8]{inputenc}
\usepackage[english]{babel}

\usepackage{forest}
\usetikzlibrary{arrows.meta}

\usepackage{authblk}
\usepackage{comment}
\usepackage{dirtytalk}
\usepackage{wrapfig}
\usepackage{multirow}
\usepackage{graphicx}
\usepackage{amsmath}

\usepackage[numbers]{natbib}

\usepackage[hyphens]{url}
\usepackage{hyperref}

\usepackage{multicol}
\usepackage{geometry}
\usepackage{fancyhdr}

\newgeometry{vmargin={15mm}, hmargin={12mm,17mm}}   

\title{Detecting East Asian Prejudice on Social Media}

\author[1,2]{Bertie Vidgen}
\author[2]{Austin Botelho}
\author[3]{David Broniatowski}
\author[1,6]{Ella Guest}
\author[1,4]{Matthew Hall}
\author[1,2]{Helen Margetts}
\author[1,3]{Rebekah Tromble}
\author[5]{Zeerak Waseem}
\author[1,2]{Scott Hale}

\affil[1]{The Alan Turing Institute}
\affil[2]{The Oxford Internet Institute}
\affil[3]{The George Washington University}
\affil[4]{The University of Surrey}
\affil[5]{University of Sheffield}
\affil[6]{The University of Manchester}

\date{May 2020}

\begin{document}

\maketitle

\begin{multicols}{2}

\paragraph{Abstract}
The outbreak of COVID-19 has transformed societies across the world as governments tackle the health, economic and social costs of the pandemic. It has also raised concerns about the spread of hateful language and prejudice online, especially hostility directed against East Asia. In this paper we report on the creation of a classifier that detects and categorizes social media posts from Twitter into four classes: Hostility against East Asia, Criticism of East Asia, Meta-discussions of East Asian prejudice and a neutral class. The classifier achieves an F1 score of 0.83 across all four classes. We provide our final model (coded in Python), as well as a new 20,000 tweet training dataset used to make the classifier, two analyses of hashtags associated with East Asian prejudice and the annotation codebook. The classifier can be implemented by other researchers, assisting with both online content moderation processes and further research into the dynamics, prevalence and impact of East Asian prejudice online during this global pandemic.\footnote{This work is a collaboration between The Alan Turing Institute and the Oxford Internet Institute.
It was funded by the Criminal Justice Theme of the Alan Turing Institute under Wave 1 of The UKRI Strategic Priorities Fund, EPSRC Grant EP/T001569/1.}


\section{Introduction}
The outbreak of COVID-19 has raised concerns about the spread of Sinophobia and other forms of East Asian prejudice across the world, with reports of online and offline abuse directed against East Asian people in the first few months of the pandemic, including physical attacks ~\cite{flanagan_canadas_2020, wong_sinophobia_nodate, liu_coronavirus_2020, walton_wuhan_nodate, solomon_coronavirus_nodate, guy_east_2020}. The United Nations High Commissioner for Human Rights has drawn attention to increased prejudice against people of East Asian background and has called on UN member states to fight such discrimination~\cite{shields_un_2020}. Thus far, most of the academic response to COVID-19 has focused on understanding its health- and economic- impacts and how these can be mitigated \cite{Latif2020}. There is a pressing need to also research and understand other forms of harm and danger which are spreading during the pandemic.

Social media is one of the most important battlegrounds in the fight against social hazards during COVID-19. As life moves increasingly online, it is crucial that social media platforms and other online spaces remain safe, accessible and free from abuse \cite{Cowls2020} -- and that people's fears and distress during this time are not exploited and social tensions stirred up. 
Computational tools, utilizing recent advances in machine learning and natural language processing, offer powerful ways of creating scalable and robust models for detecting and measuring prejudice. These, in turn, can assist with both online content moderation processes and further research into the dynamics, prevalence and impact of East Asian prejudice.

In this paper we report on the creation of a classifier to detect East Asian prejudice in social media data. It distinguishes between four primary categories: Hostility against East Asia, Criticism of East Asia, Meta-discussions of East Asian prejudice and a neutral class. The classifier achieves an F1 score of 0.83.
We also provide a new 20,000 tweet training dataset used to create the classifier, the annotation codebook and two analyses of hashtags associated with East Asian prejudice. The training dataset contains annotations for several secondary categories, including threatening language, interpersonal abuse and dehumanization, which can be used for further research.
\footnote{All research materials are available at \url{https://zenodo.org/record/3816667}.}

\section{Literature Review}
East Asian prejudice, such as Sinophobia, can be understood as fear or hatred of East Asia and East Asian people  ~\cite{bille_sinophobia_2015}. This prejudice has a long history in the West: in the 19th century the term "yellow peril" was used to refer to Chinese immigrants who were stereotyped as dirty and diseased, and considered akin to a plague ~\cite{goossen_yellow_2004}. The association of COVID-19 with China plays into these centuries old stereotypes, as shown by derogatory references to `bats' and `monkeys'~\cite{eater_2020}. Similar anti-Chinese prejudices emerged during the SARS outbreak in the early 2000s, with lasting adverse social, economic, and political impacts on East Asian diasporas globally~\cite{Lee2013}.

A 2019 Pew survey examined attitudes towards China from people in 34 countries. A median of 41\% of citizens had an unfavorable opinion of China. Negative opinions were particularly common in North America, Western Europe, and neighboring East Asian countries~\cite{silver_people_2019}. The 2019 survey, which was conducted just before the pandemic, marked a historic high in unfavorable attitudes towards China. Similarly, In 2017, a study found that 21\% of Asian Americans had received threats based on their Asian identity, and 10\% had been victims of violence~\cite{neel_poll_2017}. Likewise, a 2009 report by the Universities of Hull, Leeds and Nottingham Trent reported on the discrimination and attacks that East Asian people are subjected to in the UK, describing Sinophobia as a problem that was `hidden from public view'~\cite{Adamson2009}. Official government statistics on hate crimes are not currently available for East Asian prejudice as figures for racist attacks are not broken down by type \cite{Flatley2019}). There is relatively little research into the causal factors behind East Asian prejudice, although evidence suggests that some people may feel threatened by China's growing economic and military power~\cite{kelly_japan_2019}.

New research related to COVID-19 has already provided insight into the nature, prevalence and dynamics of East Asian prejudice, with Schild et al. demonstrating an increase in Sinophobic language on some social media platforms, such as 4chan ~\cite{schild_go_2020}. Analysis by the company Moonshot CVE also suggests that the use of anti-Chinese hashtags has increased substantially. They analysed more than 600 million tweets and found that 200,000 contained either Sinophobic hate speech or conspiracy theories, and identified a 300\% increase in hashtags that support or encourage violence against China during a single week in March 2020 \cite{TheNewStatesman2020}. East Asian prejudice has also been linked to the spread of COVID-19 health-related misinformation~\cite{cinelli_covid-19_2020}. In March 2020, the polling company YouGov found that 1 in 5 Brits believed the conspiracy theory that the coronavirus was developed in a Chinese lab~\cite{nolsoe_covid-19_nodate}. During this time of heightened tension, prejudice and misinformation could exacerbate and reinforce each other, making online spaces deeply unpleasant and potentially even dangerous.

Research into computational tools for detecting, categorizing and measuring hate speech has received substantial attention in recent years, contributing to online content moderation processes in industry, and enabling new scientific research avenues \cite{Waseem2017}. However, a systematic review of hate speech training datasets conducted by Vidgen and Derczynski shows that classifiers and training datasets for East Asian prejudice are not currently available \cite{Vidgen2020}. Somewhat similar datasets are available, pertaining to racism  \cite{Waseem2016} Islamophobia \cite{Chung2019} and `hate' in general \cite{Davidson2017, DeGibert2018} but they cannot easily be repurposed for East Asian prejudice detection. The absence of an appropriate training dataset (and, as such, automated detection tools) means that researchers have to rely instead on far cruder ways of measuring East Asian prejudice, such as searching for slurs and other pejoratives. These methods drive substantial errors \cite{Schmidt2017} as lots of less overt prejudice is missed because the content does not contain the target keywords, and non-hateful content is misclassified just because it does contain the keywords.

However, developing new detection tools is a complex and lengthy process. The field of hate speech detection sits at the intersection of social science and computer science, and is fraught with not only technical challenges but also deep-routed ethical and theoretical considerations~\cite{vidgen_challenges_2019}. Recent studies show that many existing datasets and tools contain substantial biases, such as overclassifying African American vernacular as hateful compared with Standard American \cite{Sap2019, Davidson2019}, penalising hate against certain targets more strongly than others \cite{Garg2019}, or being easily fooled by simple spelling changes that any human can identify \cite{Grondahl2018}. These issues are considerable limitations as they mean that, if used in the `real world', computational tools for hate speech detection could not only be ineffective, they could potentially perpetuate and deepen the social injustices they are designed to address. Put simply, whilst machine learning presents many exciting research opportunities, it is no panacea and tools need to be developed in dialogue with social science research if they are to be effective \cite{vidgen_challenges_2019}.

\section{Dataset Collection}
To create a 20,000 tweet training dataset, we collected tweets from Twitter's Streaming API, using 14 hashtags which relate to both East Asia and the Virus: \#chinavirus, \#wuhan, \#wuhanvirus, \#chinavirusoutbreak, \#wuhancoronavirus, \#wuhaninfluenza, \#wuhansars, \#chinacoronavirus, \#wuhan2020, \#chinaflu, \#wuhanquarantine, \#chinesepneumonia, \#coronachina and \#wohan. Some of these hashtags express anti-East Asian sentiments (e.g. '\#chinaflu') but others, such as '\#wohan' are more neutral. Data collection ran initially from 11 to 17 March 2020, and we collected 769,763 tweets, of which 96,283 were unique entries in English. To minimize biases which could emerge from collecting data over a relatively short period of time, we then collected tweets from 1st January to 10th March 2020 using Twitter's `Decahose', provided by a third party. We identified a further 63,037 unique tweets in English. The final database comprises 159,320 tweets.

We extracted the 1,000 most used hashtags from the 159,320 tweets and three annotators independently marked them up for: (1) whether they are East Asian relevant and, if so, (2) what Asian entity is discussed (e.g. China, Xi Jinping, South Korea), and (3) what the stance is towards the Asian entity (Very Negative, Negative, Neutral, Positive and Very Positive). Hashtags were also annotated for: (4) whether they are Coronavirus relevant and, if so, (5) what sentiment is expressed (Very Negative, Negative, Neutral, Positive and Very Positive). 97 hashtags were marked as either Negative or Very Negative towards East Asia by at least one annotator. All three annotations for hashtags are available in our data repository.

We then sampled 10,000 tweets at random from the database and a further 10,000 tweets which used one of the 97 anti-East Asia hashtags, thereby increasing the likelihood that prejudicial tweets would be identified and ensuring that our dataset is suitable for training a classifier \cite{Schmidt2017, vidgen_challenges_2019}.

\subsection{Data pre-processing}
We conducted one pre-processing step before presenting the tweets to annotators: hashtag replacement. Initial qualitative inspection of the dataset showed that hashtags are often used in the middle of tweets, especially when they relate to East Asia and/or Coronavirus. Hashtags which appear in the middle of tweets often play a key role in their meaning and how prejudice is expressed. For example:

\begin{quote}
its wellknown \#covid19 originated from \#china. Instead of \#DoingTheRightThing they're blaming others, typical. You cant trust these \#YellowFever to sort anything out.
\end{quote}

Without the hashtags it is difficult to discern the true meaning of this tweet, and almost impossible to be sure of whether there is any prejudice. However, in other cases, hashtags are less important to the meaning of the tweet but their inclusion could have the opposite effect -- enabling annotators to pick up on `prejudice' just because the hashtags express animosity against East Asia. This is problematic because it means that, in effect, we have returned to a dictionary based approach and any detection systems trained on such data would not generalize well to new contexts in which the same hashtags are not used. Note that this issue is usually less significant with Twitter data, where hashtags are usually only included at the end of tweets to identify broader conversations that a user wants to be part of.

To address this problem we implemented a hashtag replacement process. For the 1,000 most used hashtags, we had one annotator identify appropriate \textit{thematic replacement} hashtags. We used five thematic replacements:
\begin{itemize}
    \item{\#EASTASIA: Hashtags which relate only to an East Asian entity, e.g. \#China or \#Wuhan}
    \item{\#VIRUS: Hashtags which relate only to Corona Virus, e.g. \#coronavirus or \#covid19.}
    \item{\#EASTASIAVIRUS: Hashtags which relate to both an East Asian entity and Corona Virus, e.g. \#wuhanflu or \#chinavirus.}
    \item{\#OTHERCOUNTRYVIRUS: Hashtags which relate to both a Country (which is not East Asian) and Corona Virus, e.g. \#coronacanada or \#italycovid.}
    \item{\#HASHTAG: Hashtags which are not directly relevant to Corona Virus or East Asia, e.g. \#maga or \#spain. }
\end{itemize}

These thematic hashtag replacements were applied to all of the tweets. This means that annotators can still discern the meaning in most tweets as they are presented with the hashtags' topic but they are not unduly biased by the substantive \textit{outlook}, stance and sentiment they express. All hashtags beyond our annotated list of 1,000, were replaced with a generic replacement, \#HASHTAG. The 1,000 thematic hashtag replacements are available in our data repository.

For instance, the quote above would be transformed to:
\begin{quote}
    its wellknown \#HASHTAGVIRUS originated from \#HASHTAGEASTASIA. Instead of \#HASHTAG they're blaming others, typical. You cant trust these \#HASHTAGEASTASIAVIRUS to sort anything out.
\end{quote}

\section{Dataset Annotation}

Prejudicial language takes many forms, from overt and explicit varieties of hate, such as the use of threatening language, to subtle and covert expressions, as with microaggressions and `everyday' acts \cite{vidgen_challenges_2019}. Using a binary schema (i.e. prejudiced or not) is theoretically problematic because distinct types of behaviour, with different causes and impacts, are collapsed within one category. It can also negatively impact classification performance because of substantial within-class variation \cite{Waseem2017}. We developed the taxonomy and codebook used here iteratively, moving recursively between existing theoretical work and the data to ensure the detection system can be applied by social scientists to generate meaningful insights.

The remaining subsections outline each of the annotation categories and agreement levels. For a more detailed description see our Annotation Codebook. 

\subsection{Annotation process}
The dataset of 20,000 tweets was annotated with a three step process:

\begin{enumerate}
    \item Each tweet was initially annotated independently by two trained annotators. 
    \item Cases where annotators disagreed about the primary category were identified. Then, an Expert adjudicator made a final decision after reviewing both annotators' decisions and the original tweet (Experts were able to decide a new primary category if needed). Two experts were used, both of whom are final year PhD students working on extreme political behavior online and offline with three months experience in annotating hate speech. Having developed the annotation codebook, the two experts also had a deep understanding of the guidelines.
    \item The original annotations, where both annotators agreed, and the expert adjudications were combined to make a final dataset.
\end{enumerate}

We deployed a large team of 26 annotators, all of whom had completed at least 4 weeks of training on a previous hate speech annotation project. The final dataset of 20,000 annotations were compiled over three weeks.

\subsection{Annotations for theme}
Annotations were first made for the presence of two themes: (1) COVID-19 and (2) East Asia. This is because, despite the targeted data collection process, many of the tweets do not directly relate to either COVID-19 or East Asia. If a tweet is not identified as being East Asian relevant then no further annotations are required (it is automatically assigned to the 'neutral' class).

Annotators used an additional flag for \textit{how} they marked up the two themes, which we call `hashtag dependence'. Annotators were asked whether they identified the themes based on the text by itself or whether they had to use thematic hashtag replacements to identify them. Our account of `hashtag dependence' is very nuanced. Even in cases where hashtags are used, annotators should not rely \textit{solely} on the hashtag to identify the theme. There must be a signal in the text that the annotator picks up on, which the hashtag is then used to confirm. Archetypally, hashtag dependence relates to cases where pronouns have been used (``They", ``You", ``These"), and the link to East Asian entities is only clear once they are taken into account. For instance:

\begin{quote}
Love being out in the sun \#VIRUS \#EASTASIA
\end{quote}

This tweet would not be considered either East Asian or COVID-19 relevant because there is not a signal in the text itself which indicates the presence of the themes.

Our approach to annotating themes and the role of hashtags required substantial training for annotators (involving one-to-one onboarding sessions). This detailed annotation process means that we can provide unparalleled insight into not only what annotations were made but also \textit{how}, which we anticipate will be of use to scholars working on online communications beyond online prejudice. This initial round of annotating also helped to ensure that no tweets which were out-of-domain (i.e. not about East Asia) were accidentally annotated for categories other than Neutral.

\subsection{Primary categories}
Each tweet was assigned to one of five mutually exclusive categories -- tweets which were not marked as East Asian relevant could only be assigned to the Neutral category.

\begin{table*}[t]
\begin{center}
\begin{tabular}{ |c|c|c| } 
 \hline
 \textbf{Category} & \begin{tabular}{@{}c@{}} \textbf{Number of}\\ \textbf{Entries} \end{tabular} & \textbf{Percentage} \\
 \hline
 Hostility & 3,898 & 19.5\% \\ 
 Criticism & 1,433 & 7.2\% \\ 
 Counter speech & 116 & 0.6\% \\ 
 \begin{tabular}{@{}c@{}}Discussion of\\EA prejudice\end{tabular} & 1,029 & 5.1\% \\
 Neutral & 13,528 & 67.6\% \\
 \hline
 \textbf{TOTAL} & \textbf{20,000} & \textbf{100\%} \\
 \hline
\end{tabular}
\caption{\label{tab1}Prevalence of primary categories in the dataset.}
\end{center}
\end{table*}

\begin{itemize}
    \item \textbf{Hostility against an East Asian entity}:
    Tweets which express abuse or intense negativity against an East Asian entity, primarily by derogating or attacking them (e.g. “Those oriental devils don't care about human life” or “Chinks will bring about the downfall of western civilization”). Common ways in which East Asian hostility is articulated include: negative representations of East Asians; conspiracy theories; claiming they are a threat; expressing negative emotions. `Hostility' also includes animosity which is expressed more covertly.
    
    \item \textbf{Criticism of an East Asian entity}
    Tweets which make a negative judgment about an East Asian entity, without crossing the line into abuse. This includes questioning their response to the pandemic, how they are governed and suggesting they did not take adequate precautions and/or deploy suitable policy interventions. Note that the distinction between Hostility and Criticism partly depends upon what East Asian entity is being attacked: negativity against states tends to be criticism whilst negativity against East Asian people tends to cross into hostility.
    
    Drawing a clear line between Hostility and Criticism proved challenging for annotators. Often, Criticism would cross into Hostility when statements were framed normatively. For example, a criticism against the Chinese government (e.g. “the CCP hid information relevant to coronavirus”) could become Hostility when turned into a derogation (e.g. “It’s just like the CCP to hide information relevant to coronavirus”). However, the Hostility/Criticism distinction is crucial for addressing a core issue in online hate speech research, namely ensuring that freedom of speech is protected \cite{Ullmann2020}. The Criticism category ensures that users can engage in what has been termed `legitimate critique' \cite{Imhoff2012}, without their comments being erroneously labelled as hostile.

    \item \textbf{Counter speech}
    Tweets which explicitly challenge or condemn abuse against an East Asian entity. This includes rejecting the premise of abuse (e.g. "it isn't right to blame China!"), describing content as hateful or prejudicial (e.g. "you shouldn't say that, its derogatory") or expressing solidarity with target entities (e.g. “Stand with Chinatown against racists”).

    \item \textbf{Discussion of East Asian prejudice}
    Tweets that discuss prejudice related to East Asians but do not engage in, or counter, that prejudice (e.g. “It’s not racist to call it the Wuhan virus”). This includes content which discusses whether East Asian prejudice has increased during COVID-19, the supposed media focus on prejudice and/or free speech. Note that content which not only discussed but also expresses \textit{support} for East Asian prejudice would cross into `Hostility'.
    
    \item \textbf{Neutral}
    Tweets that do not fall into any of the other categories. Note that they could be offensive in other regards, such as expressing misogyny.
    
\end{itemize}

A small number of tweets contained more than one primary category -- but each tweet can only be assigned to one category in this taxonomy as they are mutually exclusive. To address this, we established a hierarchy of primary categories: (1) entity-directed hostility; (2) entity-directed criticism; (3) counter speech; and (4) discussions of East Asian prejudice. For example, if a single tweet contained both counter speech and hostility (e.g. “It’s not fair to blame Chinese scientists, blame their lying government”) then it was annotated as hostility.

\subsubsection{Annotation of Primary categories}
All annotations were initially completed in pairs and then any disagreements were sent to an expert adjudicator. We calculate the agreement for each pair and then take the average, minimum and maximum over all of them. Overall, agreement levels are moderate, with better results for the two most important and prevalent categories (Hostility and Neutral) but poorer performance on the less frequent and more nuanced categories, Counter Speech, Criticism and Discussion of EA prejudice. Note that if Counter Speech and Discussion of EA prejudice are combined then there is a marked improvement in overall agreement levels, with an average Kappa of 0.5 for the combined category.

\begin{table*}[t]
\begin{center}
\begin{tabular}{ |c|c|c|c| } 
 \hline
 \textbf{Measure} & \textbf{Mean} & \textbf{Min.} & \textbf{Max.} \\
 \hline
 Percentage agreement & 78\% & 67\% & 84\% \\ 
Fleiss' Kappa, All categories & 0.54 & 0.36 & 0.66 \\ 
Fleiss' Kappa, Hostility & 0.53 & 0.22 & 0.66  \\ 
Fleiss' Kappa, Criticism & 0.27 & 0.14 & 0.41 \\
Fleiss' Kappa, Counter Speech  & 0.33 & 0.11 & 0.61 \\
Fleiss' Kappa, Discussion of EA Prejudice & 0.46 & 0.14 & 0.65 \\
Fleiss' Kappa, Neutral  & 0.64 & 0.51 & 0.78 \\
\hline
\end{tabular}
\caption{\label{tab2}Agreement scores for Primary categories.}
\end{center}
\end{table*}

In the 4,478 cases (22\%) where annotators did not agree, one of two experts adjudicated. Experts generally moved tweets out of Neutral into other categories; of the 8,956 original annotations given to the 4,478 cases they adjudicated, 34\% of them were in Neutral and yet only 29\% of their adjudicated decisions were in this category. This was matched by an equivalent increase in the Hostilit category, from 31.6\% of the original annotations to 35\% of the expert adjudications. The other three categories remained broadly stable.

In 347 cases (7.7\%), experts choose a category that was not selected by either annotator. Of the 694 original annotations given to these 347 cases, 18.7\% were for Criticism compared with 39.4\% of the expert adjudications for these entries (a similar decrease can be observed for the Neutral category). Equally, the most common decision made by experts for these 347 tweets was to label a tweet as Criticism when one annotator had selected Hostility and the other selected None. This shows the fundamental ambiguity of hate speech annotation and the need for expert adjudication. With complex and often-ambiguous content even well-trained annotators can make decisions which are inappropriate.

\subsection{Secondary categories}
For Counter Speech, Discussion of East Asian prejudice and Neutral, annotators did not make secondary annotations.

For the 'Hostility' and 'Criticism' primary categories, annotators identified what East Asian entity was targeted (e.g. “Hong Kongers”, “CCP”, or “Chinese scientists”). Initially, annotators identified targets inductively, which resulted in several hundred unique targets (largely due to miss-spellings and punctuation variations). We then implemented a reconciliation process in which the number of unique targets was reduced to 78, reflecting six geographical areas (China, Korea, Japan, Taiwan, Singapore and East Asia in general).
Tweets can identify multiple East Asian targets and, where relevant, intersectional characteristics were recorded (e.g. “Chinese women” or “Asian-Americans”).

\subsubsection{Additional flags for Hostility}
For tweets which contain Hostility, annotators applied three additional flags:
\begin{itemize}
    \item Interpersonal abuse:
    East Asian prejudice which is targeted against an individual. Whether the individual is East Asian was not considered. Interpersonal abuse is a closely related but separate challenge to prejudice, and an important focus of computational abusive language research \cite{Waseem2017}.
    
    \item Use of threatening language:
    Content which makes a threat against an East Asian entity. This includes expressing a desire, or willingness, to inflict harm or advocating, supporting and inciting others to do so.
    Threats have a hugely negative impact on victims and are a key concern of legal frameworks against hate speech \cite{TheLawCommission2018, Weber2009}.
    
    \item Dehumanization
    Content which describes, compares or suggests equivalences between East Asians and non-humans or sub-humans, such as insects, weeds or actual viruses. Dehumanizing content must be literal (i.e. clearly identifiable rather than requiring a 'close reading') and must indicate malicious intent from the speaker. Dehumanizing language has been linked to real acts of genocide and is widely recognized as one of the most powerful indications of extreme prejudice \cite{LeaderMaynard2016, Musolff2015}.

\end{itemize}

The frequency of the additional flags is shown in Table 3. It shows the relatively low prevalence, even within this biased dataset, of the most extreme and overt forms of hate, with both Threatening language and Dehumanization appearing infrequently. Note that our expert adjudicators did not adjudicate for these secondary categories. In cases where experts decided a tweet is Hostile but neither of the original annotators had selected that category, none of the flags are available. In other cases, experts decided a tweet was Hostile and so only one annotation for the secondary flags is available (as the other annotator selected a different category and so did not provide any further annotations).

\begin{table*}[t]
\begin{center}
\begin{tabular}{ |c|c|c|c| } 
 \hline
 \textbf{Number of Entries} & \textbf{Threatening language} & \textbf{Dehumanization} & \textbf{Interpersonal attack} \\
 \hline
Expert Decision & 105 (2.7\%) & 105 (2.7\%) & 105 (2.7\%) \\ 
One annotation (No) & 1,347 (34.5\%) & 1,439 (36.9\%) & 1,366 (35.0\%) \\
Two annotations (No, No) & 1,989 (51.0\%) & 2,270 (58.2\%) & 2,150 (55.1\%) \\
Two annotations (No, Yes) & 251 (6.4\%) & 57 (1.5\%) & 131 (3.4\%) \\
One annotation (Yes) & 107 (2.7\%) & 15 (0.4\%) & 88 (2.3\%) \\
Two annotations (Yes, Yes) & 99 (2.5\%) & 12 (0.3\%) & 58 (1.5\%) \\
 \hline
\end{tabular}
\caption{\label{tab3}Secondary categories for Hostility.}
\end{center}
\end{table*}

\subsubsection{East Asian slurs and pejoratives}
Slurs are collective nouns, or terms closely derived from collective nouns, which are pejorative (e.g. “chinks” or “Chinazi”). Pejorative terms are derogatory references that do not need to be collective nouns (e.g. ``Yellow Fever”). Annotators marked up slurs and pejoratives as free text entry.

\subsection{Dataset availability}
The dataset described here is available in full in our Data repository. We provide it in two formats:
\begin{itemize}
    \item \textbf{Annotations Dataset}: All 40,000 annotations provided by the annotators. The only pre-processing is the provision of cleaned targets rather than the free text targets. We recommend using this dataset for investigation of annotator agreement and to explore secondary categories.
    \item \textbf{Final Dataset}: The 20,000 final entries, including the expert adjudications. We recommend using this dataset for training new classifiers for East Asian prejudice.
\end{itemize}

\section{Classification results}

Due to their low prevalence and conceptual similarity, we combined the Counter Speech category with Discussion of East Asian Prejudice for classification. As such, the classification task was to distinguish between four primary categories: Hostility, Criticism, Discussion of East Asian Prejudice and Neutral.

In order to provide a baselines for future work, we fine-tune several contextual embedding models as well as a one-hot LSTM model with a linear input layer and tanh activation. We choose the one-hot LSTM model as a contrast to the contextual embedding models. We use contextual embeddings as they take into account the context surrounding a token when generating the embedding, providing a separate embedding for each word usage. This grounding better encodes semantic information when compared with previous recurrence based deep learning approaches~\cite{vaswani2017attention}. 

The models were trained and tested using a stratified 80/10/10 training, testing and validation split. We pre-process all documents by removing URLs and usernames, lower-case the document, and replace hashtags with either a generic hashtag-token or with the appropriate thematic hashtag-token from the annotation setup. Training was conducted using the same hyper-parameter sweep identified in \cite{liu2019roberta} as the most effective for the GLUE benchmark tasks. This included testing across learning rates {1e-5, 2e-5, 3e-5} and batch sizes {32, 64} with an early stopping regime. Performance was optimized using the AdamW algorithm~\cite{loshchilov2017decoupled} and a scheduler that implements linear warmup and decay. For the LSTM baseline, we conduct a hyper-parameter search over the batch-size (16, 32, 64) and learning rate ($0.1 - 0.00001$).

\begin{table*}
\begin{center}
\begin{tabular}{|l|l|l|l|}
\hline
\textbf{Model}                                                          & \textbf{F1}            & \textbf{Recall}        & \textbf{Precision}     \\\hline
LSTM                                                           & 0.76          & 0.67          & \textbf{0.88} \\
$\text{AlBERT}_{xlarge}$ \cite{lan2019albert}  & 0.80          & 0.80          & 0.80          \\
$\text{BART}_{large}$ \cite{lewis2019bart}     & 0.81          & 0.81          & 0.83          \\
$\text{BERT}_{large}$ \cite{devlin2018bert}    & 0.82          & 0.82          & 0.83          \\
$\text{DistilBERT}_{base}$ \cite{sanh2019distilbert} & 0.80          & 0.80          & 0.81          \\
$\text{RoBERTa}_{large}$ \cite{liu2019roberta} & \textbf{0.83} & \textbf{0.83} & 0.85          \\
$\text{XLNet}_{large}$ \cite{yang2019xlnet}    & 0.80          & 0.80          & 0.82 \\\hline
\end{tabular}
\caption{\label{tab:}Classification Performance of models.}
\end{center}
\end{table*}

All of the contextual embedding models outperformed the baseline to varying degrees. RoBERTa achieved the highest F1 score (0.83) of the tested models, which is a 7-point improvement over the LSTM (0.76). This model harnesses the underlying bidirectional transformer architecture of \cite{devlin2018bert}, but alters the training hyperparameters and objectives to improve performance. Curiously, we see that the LSTM baseline outperforms all other models in terms of precision.

For the best performing model (RoBERTa), sources of misclassification are shown in the confusion matrix. This shows that the model performs well across all categories, with strongest performance in detecting tweets in the Neutral category (Recall of 91.6\% and Precision of 93\%). The model has few misclassifications between the most conceptually distinct categories (i.e. Hostility versus Neutral or Discussion of East Asian Prejudice) but has far more errors when distinguishing between the conceptually more similar categories, Criticism and Hostility.

\begin{figure*}
\begin{center}
\includegraphics[width=9cm]{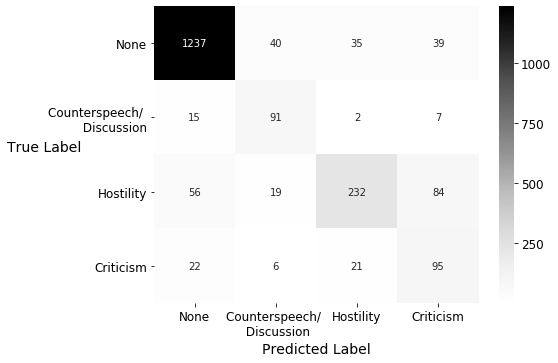}
\caption{\label{fig:1}Confusion Matrix for RoBERTa Classifications.}
\end{center}
\end{figure*}

\section{Error analysis}

To better understand not only the extent of misclassification but also the sources of error, we conducted a qualitative analysis of misclassified content, using a grounded theory approach \cite{Corbin1990}. This methodology for qualitative research is entirely inductive and data-driven. It involves systematically exploring themes as they emerge from the data and organizing them into a taxonomy – refining and collapsing the categories until 'saturation' is reached and the data fits neatly into a set of mutually exclusive and collectively exhaustive categories. Figure 1 shows the error categories within our sample of 340 misclassified tweets from the 2,000 (10\%) validation split. The errors broadly fit within two branches, annotator errors (17\%) and machine learning errors (83\%).

\subsection{Annotator Errors, (17\% of total)}
Annotator errors are cases where the class predicted by the model may better represent the tweets' content and be more consistent with our taxonomy and guidelines. In effect, we believe that the 'wrong' classification provided by the model is right -- and that a mistake may have been made in the annotation process. Approximately 17\% (N=58) of the errors were due to this. Note that this does not mean that 17\% of the dataset is incorrectly annotated as this sample is biased by the fact that it has been selected precisely because the model made an 'error'. The actual prevalence of misannotated data is likely far lower.  

36 of the annotator errors were clear misapplications of primary categories. The other 22 were cases where annotators made detailed annotations for tweets which were not East Asian relevant. Often, this was because annotators over-relied on hashtags for discerning East Asian relevance. These are \textit{path dependency errors} and show the importance of annotators following the right instructions throughout the process. If a misannotation is made early on then the subsequent annotations are likely to be flawed.

\subsection{Machine Learning Errors, (83\% of total)}
83\% of the total errors were due to errors from the model. We have separated these errors into edge and non-edge cases. Non-edge cases are where the model has made an error that is easily identified by humans (comprising 37\% of all errors); edge-cases are where the misclassified content contains some ambiguity and the model misclassification has some merit (comprising 46\% of all errors).

\subsubsection{Edge Cases}
\paragraph{Hostility vs. Criticism}
Misclassifying Hostility as Criticism (and vice versa) was the largest source of error, reflecting also the high levels of annotator disagreement between these categories. The model struggled with cases where criticism was framed in a normative way (e.g. “gee, china was lying to us. what a bloody shock”). In such cases, the model misclassified the tweets as Criticism rather than Hostility.

\paragraph{Discussion of East Asian prejudice vs. Neutral}
The model misclassified Neutral as Discussion of East Asian prejudice in several tweets. These were usually cases where the virus was named and discussed but \textit{prejudice} was not discussed explicitly (e.g. “corona shows why you should blame all problems with China on Trump”).

\paragraph{Co-present Primary Categories}
In our taxonomy, annotators could only assign each tweet to a single primary category. However, in some cases this was problematic and the model identified a co-present category rather than the primary category which had been annotated. For instance, the model identified the following tweet as 'Discussion of East Asian Prejudice': “don't sell your mother for chinese yuan. stop being HASHTAGEASTASIA propaganda tool. calling HASHTAGEASTASIA+VIRUS where it came from, isn't racist but truth”. It missed the criticism of China, which was also co-present and for which it had been annotated. 

\paragraph{Ambiguous Content}
In some cases, the content of tweets was ambiguous, such as positively framed criticism (e.g. “china official finally admits the HASHTAGEASTASIA+VIRUS outbreaks”) or use of grammatically incorrect language.

\subsubsection{Non-edge Cases}
\paragraph{Identification Errors}
In several cases the misclassified tweets had clear textual signals which indicate why the tweets had been misclassified, such as the presence of signal words and phrases (e.g. 'Made in China'). This is a learned over-sensitivity to signals which are frequently associated with each primary category.

\paragraph{Target Errors}
In over a third of all non-edge case errors the model identified an appropriate category -- but the target was not East Asian and so the classification was wrong. For instance, tweets which expressed hostility against an invalid target (e.g. India, WHO or the American government) were routinely classified as Hostility. In such cases, an appropriate entity (i.e. China) was usually referred to but was not the object of the tweet, creating a mixed signal. Conversely, in other cases, the model failed to identify Criticism or Hostility against an East Asian entity because it required some context-specific cultural knowledge (e.g. “building firewall and great wall is the only thing government good at! HASHTAGVIRUS HASHTAGEASTASIA+VIRUS”). East Asian prejudice that targeted a well-known East Asian person, such as Xi Jinping, was also often missed.

\paragraph{Errors due to Tone}
In some tweets, complex forms of expression were used, such as innuendo or sarcasm (e.g. “I think we owe you china, please accept our apologies to bring some virus into your great country”). Although the meaning is still discernible to a human reader, the model missed the important role played by tone.

\subsection{Addressing Classification Errors}
Classifying social media data is notoriously difficult with many different sources of error which, in turn, require many potential remedies. The prevalence of annotator errors, for example, illustrates the need for robust annotation processes and providing more support and training when applying taxonomies. Removing such errors entirely is arguably impossible, but better annotation processes and guidelines would help to limit the number of avoidable misclassifications.

Edge cases are a particularly difficult type of content for classification, and there is not an easy solution to how they should be handled. Edge cases can be expected in any taxonomy that draws distinct lines between complex and non-mutually exclusive concepts, such as hateful speech and legitimate criticism. Nonetheless, larger and more balanced datasets (with more instances of content in each category) would help in reducing this source of error. Equally, the frequency of non-edge case machine learning errors (i.e. cases where the model made a very obvious mistake) could also be addressed by larger datasets, as well as more advanced machine learning architectures.

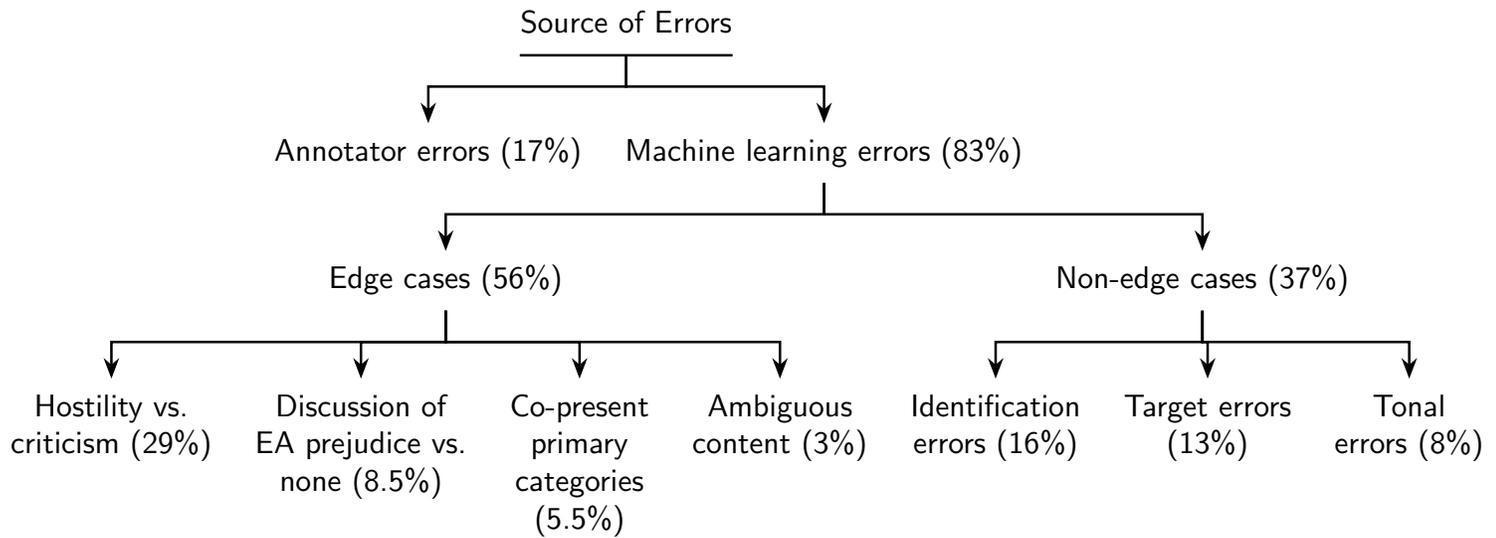
\begin{figure*}
\begin{center}
\resizebox{20cm}{!}{\begin{forest}
  for tree={
    align=center,
    parent anchor=south,
    child anchor=north,
    font=\sffamily,
    edge={thick, -{Stealth[]}},
    l sep+=10pt,
    edge path={
      \noexpand\path [draw, \forestoption{edge}] (!u.parent anchor) -- +(0,-10pt) -| (.child anchor)\forestoption{edge label};
    },
    if level=0{
      inner xsep=0pt,
      tikz={\draw [thick] (.south east) -- (.south west);}
    }{}
  }
  [Source of Errors
    [Annotator errors (17\%)]
    [Machine learning errors (83\%)
      [Edge cases (56\%)
      [Hostility vs.\\criticism (29\%)]
      [Discussion of\\EA prejudice vs.\\none (8.5\%)]
      [Co-present\\primary\\categories\\(5.5\%)]
      [Ambiguous\\content (3\%)]
      ]
      [Non-edge cases (37\%)
        [Identification\\errors (16\%)]
        [Target errors\\(13\%)]
        [Tonal\\errors (8\%)]
      ]
    ]
  ]
\end{forest}}
\caption{\label{fig:2}Sources of Classification error.}
\end{center}
\end{figure*}

\section{Hashtag analysis}

Because annotators were presented with tweets where all hashtags were replaced with either a thematic hashtag replacement or just 'hashtag' as a placeholder, we can conduct unbiased analyses on the co-occurrence of hashtags with the annotated primary categories. For each category, we filtered the data so only hashtags which appeared in at least ten tweets assigned to that category were included. Then, we ranked the hashtags by the percentage of their uses which were accounted for by the primary category. For brevity, only the twenty hashtags most closely associated with the Hostility category are shown here. 

A small number of hashtags are highly likely to only appear in tweets that express Hostility against East Asian entities. These hashtags can be used to identify prejudiced discourses online and, in some cases, their uses may indicate intrinsic prejudice. Surprisingly, many seemingly hostile hashtags against East Asia, such as \#fuckchina and \#blamechina are not always associated with hostile tweets (in these cases, Hostility accounts for 67.5\% and 60.5\% of their total use, respectively). This shows the importance of having a purpose-built machine learning classifier for detecting East Asian hostility, rather than relying on hashtags and keywords alone.

\begin{table*}
\begin{center}
\begin{tabular}{ |c|c|c|c| } 
 \hline
 \textbf{Hashtag} & \begin{tabular}{@{}c@{}}\textbf{Frequency in}\\\textbf{Hostile tweets}\end{tabular} & \begin{tabular}{@{}c@{}}\textbf{Percentage of}\\\textbf{all uses}\end{tabular} & \begin{tabular}{@{}c@{}}\textbf{Number of}\\\textbf{total uses}\end{tabular} \\
 \hline
 \#rule2 & 20 & 87.0\% & 23\\ 
 \#rule3 & 17 & 85.0\% & 20\\ 
 \#rule1 & 22 & 84.6\% & 26\\ 
 \#makechinapay & 18 & 72.0\% & 25\\
 \#hkgovt & 22 & 71.0\% & 31\\
 \#fuckchina & 54 & 67.5\% & 80\\
 \#blamechina & 23 & 60.5\% & 38\\
 \#batsoup & 15 & 60.0\% & 25\\
 \#hkairport & 11 & 55.0\% & 20\\
 \#huawei & 16 & 53.3\% & 30\\
 \#boycottchina & 185 & 52.9\% & 350\\
 \#communismkills & 14 & 51.9\% & 27\\
 \#communistchina & 34 & 50.7\% & 67\\
 \#chinaisasshoe & 41 & 50.6\% & 81\\
 \#chinapropaganda & 10 & 50.0\% & 20\\
 \#china\_is\_terrorist & 168 & 48.7\% & 345\\
 \#xijingping & 17 & 48.6\% & 35\\
 \#chinashouldapologize & 14 & 48.3\% & 29\\
 \#madeinchina & 39 & 47.6\% & 82\\
 \#ccp & 395 & 46.5\% & 850\\
 \hline
\end{tabular}
\caption{\label{tab4}Hashtags in Hostile tweets.}
\end{center}
\end{table*}

\section{Conclusion}
East Asian prejudice is a deeply concerning problem linked with COVID-19, reflecting the huge social costs that the pandemic has inflicted on society, as well as the health-related costs. In this paper we have reported on development of several research artefacts which we hope will enable future research into this source of online harm. We make these available in our repository:

\begin{enumerate}
    \item A classifier for detecting East Asian prejudice.
    \item A 20,000 tweet training dataset used to create the East Asian prejudice classifier.
    \item A 40,000 annotations training dataset, which contains the full annotations made by each annotator.
    \item A list of hashtag `replacements', where COVID-19 specific hashtags are swapped out with thematic replacements.
    \item Three sets of annotations for the 1,000 most used hashtags in the original database of COVID-19 related tweets. Hashtags were annotated for COVID-19 relevance, East Asian relevance and stance.
    \item The full codebook used to annotate the tweets. This contains extensive guidelines, information and examples for hate speech annotation.
\end{enumerate}

\bibliographystyle{unsrtnat}
\bibliography{lit-review.bib}


\emergencystretch=1em

\end{multicols}

\end{document}